\documentclass[conference,a4paper]{ieeetran}

\usepackage[latin1]{inputenc}
\usepackage{subeqnarray}
\usepackage{amsfonts}
\usepackage{amsmath}
\usepackage{amssymb}
\usepackage{graphicx}
\usepackage{color}
\usepackage{array}
\usepackage{balance}
\title{Local feature hierarchy for face recognition across pose and illumination}

\author{\authorblockN{Xiaoyue Jiang, Dong Zhang and Xiaoyi Feng }
\authorblockA{ Northwestern Polytechnical University \\
e-mail: xjiang@nwpu.edu.cn}
}

\begin{document}
\maketitle
\begin{abstract}
Even though face recognition in frontal view and normal lighting condition works very well, the performance degenerates sharply in extreme conditions. Recently there are many work dealing with pose and illumination problems, respectively. However both the lighting and pose variation will always be encountered at the same time. Accordingly we propose an end-to-end face recognition method to deal with pose and illumination simultaneously based on convolutional networks where the discriminative nonlinear features that are invariant to pose and illumination are extracted. Normally the global structure for images taken in different views is quite diverse. Therefore we propose to use the $1\times1$ convolutional kernel to extract the local features. Furthermore the parallel multi-stream multi-layer $1\times1$ convolution network is developed to extract multi-hierarchy features. In the experiments we obtained the average face recognition rate of $96.9\%$ on multiPIE dataset,which improves the state-of-the-art of face recognition across poses and illumination by $7.5\%$. Especially for profile-wise positions, the average recognition rate of our proposed network is $97.8\%$, which increases the state-of-the-art recognition rate by $19\%$.

\end{abstract}

\begin{keywords}
   face recognition, pose variation, illumination variation, local feature, convolutional neural networks.
\end{keywords}

\section{Introduction}
Face recognition has been one of the most active research topics in computer vision for more than three decades. With years of efforts, promising results have been achieved for automatic face recognition in both controlled\cite{TIP10Tan} and uncontrolled environments \cite{PAMI16Ding,Gunther2013}. A number of algorithms have been developed for face recognition with wide variations in view and illumination, respectively. Yet few attempts have been made to tackle the face recognition problem with the variations of both pose and illumination \cite{ACM03Zhao}. In fact, face recognition remains significantly affected by both pose and illumination which are often encountered in real world images. Recognizing faces reliably across pose and illumination has been proved to be a much more difficult problem.

Pose variation induces dramatic appearance changes in the face image. Essentially, this is caused by the complex 3D geometrical structure of the human head. The rigid rotation of the head results in self-occlusion which means that some facial appearance will be invisible. At the same time the shape and position of the visible part of facial image also vary nonlinearly from pose to pose. Consequently, the appearance diversity caused by pose is usually greater than that caused by identity. Thus the general face recognition algorithms always fail when dealing with the images of different poses.

Illumination will also cause dramatic variations for images.  Assuming Lambertian reflectance, the intensity value $I(x, y) $of every pixel in an image is the product of the incident lighting $L(x, y)$ and the reflectance $R(x, y)$ at that point as $I(x, y) = R(x, y) \times  L(x, y)$. Thus, the captured images varies with the incident lighting. In order to achieve face recognition across illumination, there are two kinds of strategies. One is to extract illumination invariant features from images, such as LBP \cite{PAMI06LBP}, HOG \cite{ICCV09HOG} et al., the other is to model the distribution of illumination \cite{14:Lee:05, ECCV08_Jiang}.

In real applications, both the pose and illumination variations exist. Thus a robust face recognition system should be able to deal with the both variations at the same time. Recently, the deep learning methods \cite{NIPS14Zhu,CVPR16Kan} showed its great ability to model nonlinear distributions of data. It achieved the state-of-the-art performance in many fields of pattern recognition, such as object classification \cite{CVPR15Szegedy}, object detection \cite{NIPS15FRCNN}.  Its great capacity is mainly due to the learning procedure which can find the hierarchical features from dataset. These features from each layer of the network include the whole hierarchy of the objects from a local gradient to its global structure. As a result, these learned features are more informative than traditional human engineered features.

Even though different poses can induce the different appearance of the face, there are some correlations between images of the same identity in different poses. Similarly, images of the same identity in different illumination also correlate to each. Thus through a proper learning method, the pose and illumination invariant features can be obtained. Inspired by the excellent feature learning ability of deep convolutional networks, it is employed to develop an end-to-end face recognition method across pose and illumination in this paper.

\begin{figure}
\label{fig:netStruc}
\centering
\includegraphics[height=10cm]{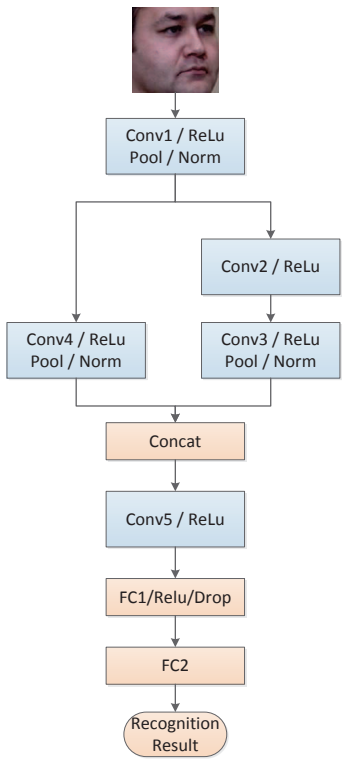} 
\caption{Architecture of the proposed deep network. Conv1 has the kernel size of $11\times11$ and the dimension of 96. Conv2, Conv3, Conv4 and Conv5 all have the kernel size of $1\times1$, and the dimension of 200, 400, 300 and 500 ,respectively.}
\end{figure}

The traditional ConvNet is good at extracting hierarchical features, however these features are embedded in the networks. Then it always becomes too late for the final softmax layer to do the recognition. In fact, multi-scale information is also very important for many tasks \cite{ICCV15Xie}. Therefore we introduce a multi-stream architecture to extract multi-scale features instead of the traditional cascade one-stream structure, as shown in Fig.\ref{fig:netStruc}. Within each stream the features of different hierarchy level are extracted. And then they are combined for classification in the final stage. For face recognition, especially across illumination and pose, local features contribute more than global features. Thus $1\times1$ convolutional filters are introduced to abstract local features.




The key contributions of our work are (i) we propose an end-to-end network for face recognition across pose and illumination. Images from different views are all go into a same input of the network. The proposed network is more compact than multi-input network such as \cite{CVPR16Kan}. (ii) A parallel structure is proposed to extract multi-hierarchy features from the root layer. In order to extract local features of images, only $1\times1$ convolutional kernel is applied. Then with multiple layers of $1\times1$ convolutional kernel, the highly discriminative local nonlinear features are extracted. (iii) The proposed network is not very deep, thus the number of the parameters is relatively small and easy to be trained. We demonstrate the effectiveness of the proposed method by facial images over different poses and illumination in the controlled environment (Multi-PIE).

\section{Related Work}

Recently, many promising approaches have been proposed to tackle the poses and illumination challenge in face recognition. For the problem of pose, they are either handled by 2D image matching or by encoding a test image using some bases or exemplars. Carlos et al. \cite{CVPR11Castillo} used stereo matching to compute the similarity between two faces. Li et al. \cite{TIP12Li} proposed a test face as a linear combination of training images, and utilized the linear regression coefficients as features for face recognition. Tied Factor Analysis \cite{PAMI08Prince} tried to find the identity feature regardless of poses through a group of angle specified linear functions by a probabilistic approach. Ding et al. \cite{TIP15Ding} first transformed the non-frontal faces to frontal faces, then they applied patch-based partial representation (PBPR) to describe the transformed frontal face. Also there are some methods tried to find a suitable metric for images across pose. Kan et al.\cite{PAMI16Kan} proposed a Multi-view Discriminant Analysis (MvDA) approach, which seeks for a single discriminant common space for multiple views by jointly learning multiple view-specific linear transforms. Due to the great performance of ConvNet, Zhu et al. \cite{ICCV13Zhu}\cite{NIPS14Zhu} used ConvNet to recover frontal images for images under arbitrary views. Kan et al.\cite{CVPR16Kan} proposed a multi-view deep network for cross-view face recognition, which has several different inputs for images of different views respectively.

For illumination, current attempts to handle this variation is either to find invariant features and representations or to model this variation. Algorithms based on Retinex theory \cite{TIP10Tan} \cite{chenPAMI06} tried to separate lighting from illumination with different prior of lighting or reflectance, thus an illumination invariant facial representation can be used for further recognition.  The other kind of work is try to model the illumination based on the observation of illumination cone theory, including spherical harmonic subspace estimation \cite{9:Ramamoorthi:01}, 9-point light subspace estimation \cite{14:Lee:05}, or even recover lighting subspace from an arbitrary pose \cite{ECCV08_Jiang} and so on. Sparse representation also showed its effectiveness in dealing with lighting problems \cite{PAMI12Wagner}

Besides to deal with pose and illumination problem separately, there are some research tried to handle the combined variation of pose and illumination. 
There are two categories of method to deal with this two factors altogether. One is to consider the illumination problem and pose problem as two independent problems. Then the recognition problem is solved by using methods dealing with illumination and poses one after another \cite{JainHandbook}. The second category of methods to deal with these two kinds of variation is to recover the canonical facial images. Sparse coding method \cite{PAMI12Wagner} and deep learning-based method \cite{ICCV13Zhu} are employed to encode faces in arbitrary views and lighting conditions.

In this paper, our target is to recognize each face, so the identity of each face is used as the output of the networks but not the recovered canonical face image where a further recognition method, such as SVM, is need to identify the face. The merit of using identity as output for recognition task is in two aspects. First it is an end-to-end structure for the recognition task. Secondly using an identity as the output of the network is much easier to be convergent than using face images as output in training. That is because the number of parameters is much smaller for the classification task than that for the regression task. In order to achieve the face recognition across pose and illumination, the local non-linear pose and illumination invariant representations should be extracted. In the proposed networks, the $1\times1$ convolutional kernel is used to extract local texture. While the parallel multi-layer network structure of network is to extract the multi-hierarchy nonlinear local features for recognition.

\section{Multi-Stream Convolutional Neural Networks}

For face recognition across pose and illumination, the global structure of images is destructed by views, meanwhile, lighting brings wide variations for the appearance of images. Thus the discriminative features invariant to pose and illumination should be local but not global features. Furthermore the multiple hierarchical features are always much more informative than features in a single scale. Consequently, we propose an end-to-end convolutional network which can extract multi-hierarchy  local features for the task of face recognition. The overall architecture is shown in Fig.1. In our proposed networks, the input is a facial image under an arbitrary pose and illumination. The output is the identity label for the face image.

\subsection{Root convolutional layer}

Recently convolutional neural networks (CNN) showed great performance in different fields of computer vision, such as object detection \cite{NIPS15FRCNN} and object classification or recognition \cite{CVPR15Szegedy}. The superb capability of CNN is mainly due to its high order nonlinear representation for data. In practice, CNN extract features from input images layer by layer through convolution kernels.
For the proposed networks, the input images, $x_i^{l-1}$, are cropped and mirrored to the size of $w\times h\times c=227 \times 227 \times 3$. Then they are fed into a convolutional layer (Conv1) $k_{ij}^{l}$ with 96 filters of size $11\times 11\times3$. The output ,$x_j^l$, of this convolutional layer is written as,

\begin{equation}\label{eq:Conv}
    x_j^l = \sum_{i \in {M_j}}x_i^{l - 1}*k_{ij}^l + b_j^l
\end{equation}
where $l$ is the layer index. $b_j^l$ is the additive bias term. $*$ represents convolution in a local selection $M_j$ of input signals. In this convolutional layer, 96 filters are applied locally to the whole images resulting a feature map of size $55\times55\times96$. Then rectified linear unit (ReLU) is applied to the extracted feature map. ReLU serves as an activation unit in the network which brings the non-linearity to the feature. Here we use a ramp function $f(x)=max(0,x)$ to rectify the feature map. This activation function is considered to be more biologically plausible than the widely used logistic sigmoid or hyperbolic tangent function.

Consequently, the rectified feature maps will be given to a max-pooling layer (Pool) which takes the max over $3\times3$ spatial neighborhoods with a stride of 2 for each channel, respectively. Through max-pooling operation, features will become insensitive to local shift, i.e. invariant to location. Afterwards, those features will go through the local response normalization layer (Norm), which performs the lateral inhibition by normalizing over local input regions. In our model, the local regions extend across nearby channels, but have no spatial extent. For normalization, each input value is divided by the sum of local region as shown in Eq.\ref{eq:localSum}
\begin{equation}
\label{eq:localSum}
    s(x_i)=(k+(\alpha/n)\sum_i x_i^2)^\beta
\end{equation}
where $n$ is the size of each local region, and the sum is taken over the region centered at that value $x_i$ (zero padding is added where necessary). From the root layer, we can obtain the local feature set which mainly contains all kinds of edges in different orientations. Generally edges are considered as illumination invariant features. Since local structures are more important in our case, we will continue to seek for local features instead of global features which are normally obtained in further layers of traditional CNN.

\subsection{multi-hierarchical local feature}
In fact, the window size of the convolution kernel is considered as the receptive field for feature extraction. That is bigger windows can include information in wider range for processing. For the face images from different views, the global structure of images changes diversely. However, there is a tight correlation among local regions of images taken in different views. Therefore the pose-invariant features should be local features, and in addition the spatial information should be kept for each local feature. Accordingly, smaller windows should be applied to extract features. Here we propose to use the kernel of size $1 \times1 \times c$. With the kernel of $1\times1$, no spatial patterns across multiple pixels are extract, but the patterns between $c$ channels are learned without losing the location information for each pattern. Thus the feature can keep the correlation among different views. Meanwhile, the number of parameters will also be reduced for the $1\times1$ kernel size compared with that of bigger kernel size, which can make the training procedure more easily to be convergent.

Neural networks actually perform non-linear operation for data. With multiple layers of processing, the network can build a high-order non-linear model for the real images, which is a much more suitable representation for the data than the traditional man-crafted features. As a result, different numbers of layers also influence the property of features. In the classical ConvNet, there is only one path for the signal to go, and the classification only performs on the features extracted by the last layer of the network. In fact, features from different level of the networks all contain useful information. Thus we propose to build a multi-stream local feature hierarchy network (LFHN). Within each stream, the features of different order are extracted by using different number of convolutional layers. Then features from different stream are concatenate to compose a multi-hierarchy feature with the size of $h\times w \times (c_1+c_2+\cdots+c_n)$, where $ h\times w \times c_i$ is the size of the feature from stream $i$, and $c_i$ is the depth dimension.

 \begin{table*}[t]
 \centering
 \caption{Rank-1 identification rates on combined variations of pose and illumination on MultiPIE.}
  \begin{tabular}{|p{0.8cm}||m{0.6cm}|m{0.6cm}|m{0.6cm}|m{0.6cm}|m{0.6cm}|m{0.6cm}|m{0.6cm}|m{0.6cm}|m{0.6cm}|m{0.6cm}|m{0.6cm}|m{0.6cm}|m{0.6cm}|m{0.6cm}||m{0.6cm}|}
  \hline
  PoseID & 081 & 110 & 120& 090 & 080 & 130 & 140 & 050 & 041 & 190 & 200 & 010 & 240 & 191& \\
  \hline
  Yaw&  $-45^o$  & $-90^o$ & $-75^o$ & $-60^o$ & $-45^o$ & $-30^o$ & $-15^o$ & $15^o$ & $30^o$ & $45^o$  & $60^o$ & $75^o$ & $90^o$ & $45^o$ &\\
  Pitch & $25^o$ & $0^o$   & $0^o$ & $0^o$ & $0^o$ & $0^o$ & $0^o$ & $0^o$ & $0^o$ & $0^o$ & $0^o$  & $0^o$ & $0^o$  & $25^o$ & Mean \\
  \hline
  \hline
  RR
  \cite{TIP12Li}&24     & 20.5	&26.5&	50.64&65.30&70.97 &	81.07&	77.21	&73.69&	58.12	&45.97&	31	&18&40& 48.78 \\
  \hline
  FIP
  \cite{ICCV13Zhu} & -  & -  & - & -&67.10&	74.60&	86.10&	83.30&	75.30&	61.80 & -  & -  & - & -&-\\
  \hline
  PBPR
  \cite{TIP15Ding} & 88	&51	&79	&90.86&	\textbf{97.91}&	\textbf{99.41}&	\textbf{99.05}&	\textbf{99.94}&	 \textbf{99.23}&	\textbf{98.21}&	87.75&	89&	75&	96 &89.31 \\
  \hline
  LFHN
  (Ours) & \textbf{94.51}	&\textbf{97.52}	&\textbf{98.15}	&\textbf{98.51}	&97.75&	98.08&	97.12&	93.74&	97.91&	 96.53&	\textbf{97.65}&	\textbf{97.57}&	\textbf{97.3}&	\textbf{93.73} & \textbf{96.86}\\
  \hline
  \end{tabular}

  \label{tab:PI}
\end{table*}

In order to keep the spatial information for the local features, the convolutional kernel of $1\times1$ is applied. As shown in Fig.\ref{fig:netStruc}, there are two steams for the proposed network. One stream contains two convolutional layers where the kernel size of Conv2 and Conv3 is $96\times1\times1\times200$ and $200\times1\times1\times400$, respectively. In another stream, there is only one convolutional layer Conv4 with the kernel of $96\times1\times1\times300$. Through these two streams, we can get local features from different levels of hierarchy. In order to achieve the final recognition task, one more convolution layer, Conv5 with kernel size $700\times1\times1\times500$, and two fully connected layer are employed for the further feature abstraction.

\subsection{training}

Since the root layers of convolutional networks always contain more generic features such as edges or color blob, which is useful for many tasks including face recognition. In training, we keep the pretrained results of AlexNet as the weight for the root convolutional layer, Conv1. Then for the other layers, they are trained according to the Softmax loss function based on the identity labels for images in MultiPIE dataset \cite{MultiPIE}.

\section{Experiments}
\subsection {Dataset}

To evaluate the effectiveness of the proposed local feature hierarchy networks (LFHN) under different poses and illumination, the MultiPIE face database \cite{MultiPIE} was employed.  The MultiPIE face database contains 754,204 images of 337 identities. Each identity has images captured under 15 different poses and 20 different lighting conditions. For the original images in MultiPIE, we have aligned all the images according to the position of eyes and crop them to the size of $256\times256$. For each subject, we only select the images with neural expression but in all poses and lighting conditions, thus for each person there are 300 images. Altogether, for all the individuals in the dataset, we put $300 \times 337 = 101100$ images into the data pool for training and testing. In MultiPIE, there are 4 sessions to take photos for each subject, but not everyone comes in each session. Therefore we take all the images of 250 individuals in session 1, and the images of the other 87 persons in session 2 and 3. 

\subsection{recognition across poses and illumination}
In order to evaluate the performance of the proposed local hierarchy networks, MultiPIE is used to train and test the networks. For the proposed networks, there is only one input instead of multiple input for images from different reviews. Therefore training images are randomly selected from images with natural expression but in 15 different poses and 20 different lighting conditions. For all the 337 individuals, 90000 images are randomly taken from 101100 images for training, and the leftover images are used as testing images. The proposed network can learn local non-linear features which can represent the correlations between images in different poses and lighting conditions. The rank-1 recognition rates for images with pose and illumination variations are shown in Table.\ref{tab:PI}, where the bold numbers are the best results. The recognition results for each view is the average results for all the images under 20 different lighting conditions.


From the results, we can see that the proposed LFHN network achieved relatively stable performance for different poses. Especially for profile-wise images, where the yaw angle is in the range of $[-90^o,-60^o]$ and $[60^o, 90^o]$, the average recognition rate is $97.8\%$. While for the patch-based partial recogniton(PBPR)\cite{TIP15Ding}, the performance declined significantly to $78.8\%$ for profile-wise images. For PBPR, the average recognition rate for front-wise images where the yaw angle is within $\pm45^o$ is $97.2\%$. That indicates the performance of the projection recovery method used in PBPR degenerated when the pose variation becomes greater. Compared with current state-of-the-art methods, our proposed LFHN network improves the recognition rate by $7.55\%$ for images under arbitrary poses and illumination, especially for profile-wise images the recognition rate increases by $19\%$. In the proposed networks, we consider images from different views and illumination equally and do not try to re-project images to frontal view. Thus pose-invariant and illumination-invariant nonlinear local features can be directly seek by the proposed network LFHN. 

\begin{table*}[t]
  \centering
  \caption{Rank-1 identification rates for pose variation on MultiPIE.}
  \label{tab:P2}
   \begin{tabular}{|p{1.8cm}||c|c|c|c|c|c|c|c|c|c|c|c||c|}
   \hline
   PoseID & 110 & 120& 090 & 080 & 130 & 140 & 050 & 041 & 190 & 200 & 010 & 240&  \\
   Yaw &$-90^o$ &	$-75^o$ & $-60^o$ &	$-45^o$ &$-30^o$ &$-15^o$ &	$15^o$ & $30^o$ & $45^o$ & $60^o$ &	$75^o$ &$90^o$ &Mean\\
   \hline
   \hline
   PLS \cite{CVPR11Sharma}&	31.9&	77.5&	89.2&	93.4&	88.3&	98.1&	98.1&	93.4&	90.6&	87.3&	72.3&	 26.8&	78.9\\
   \hline
   MCCA \cite{KDD10Rupnik} &	40.9&	74.2&	82.2&	72.3&	68.5&	92.0&	90.6&	79.8&	74.7&	77.9&	71.4&	 37.6&	71.8\\
   \hline
   PLS+LDA \cite{CVPR16Kan}&  38.0&	79.8&	86.9&	94.4&	92.0&	99.5&	98.6&	96.7&	88.3&	85.0&	70.9&	 31.9&	80.2\\
    \hline
   MCCA+LDA \cite{CVPR16Kan} &	48.8&	 66.2&	 81.7&	 88.7&	100&	100&	100&	 99.5&	 83.1&	80.3&	 67.6&	 56.8&	  81.1\\
    \hline
    MvDA \cite{PAMI16Kan}  &	 56.8&	 72.3&	 84.5&	 92.0&	 96.7&	100&	100&	 99.1&	 89.7&	86.4&	 71.4&	 55.9&	  83.7\\
    \hline
    GMA \cite{CVPR12Sharma} &	 52.6&	 73.2&	 84.5&	 90.1&	100&	100&	100&	100&	 90.6&	85.9&	 71.8&	 57.3&	  83.8\\
    \hline
    MvDN \cite{CVPR16Kan}& 	 70.4&	 82.2&	 88.3&	 91.1&	\textbf{ 99.1}&	\textbf{100}&	\textbf{100}&	\textbf{ 99.1}&	  93.0&	 91.1&	 79.8&	 70.9&	 88.7\\
    \hline
    LFHN(ours)&	\textbf{100}&	\textbf{ 97.7}&\textbf{100}&\textbf{100}&	 89.3 &	100&	 93.8&	97.1 &	\textbf{ 95.8} &	 \textbf{ 93.5}& 	\textbf{100} &	\textbf{100}&	\textbf{ 97.3}\\
   \hline
 \end{tabular}
\end{table*}

Besides the face recognition with the combined variations of pose and illumination, we also test the performance of the proposed network LFHN on pose only. For this task, the probe dataset includes images of all subjects from 4 sessions where only the images taken in ambient lighting and 13 different poses are selected. The comparison results with other methods is shown in Table \ref{tab:P2}, where the bold numbers are the best results.


From the results we can see even though different methods tried to tackle the pose problem, the face recognition rate decreased along with the amount of pose variation. The more the view diverged from the frontal face, the lower recognition rate is. That is because pose changes the appearance and structure of the image. MvDN \cite{CVPR16Kan} and MvDA \cite{PAMI16Kan} tried to find the correlation between different poses, and achieved relatively good performance. For the front-wise images where yaw angle is within $\pm45^o$, the mean recognition rate for MvDN is $97.1\%$, while for the profile-wise images where yaw angle is bigger than $\pm45^o$, the mean recognition rate for MvDN declines to $80.5\%$. For the proposed network LFHN, we focus on the extraction of local features among different poses which can describe the correlation of different poses and also discriminate different identity. Thus we achieved better results compared with other methods. Especially for larger pose diversity, the performance of the proposed network is not degenerated but very stable instead. The average recognition rate for all different poses is $97.3\%$, which improves the state-of-the-art method by $8.6\%$. For the front-wise images,the mean recognition rate is $96.0\%$, which is comparative to the result of MvDN. While for the profile-wise images, the mean recognition rate is still $98.5\%$ which improves the state-of-the-art method MvDN by $18\%$.


\section{Conclusion}
Pose and illumination will always bring great variance for the appearance of face images, which makes face recognition across pose and illumination challenged. However, it is quite normal to encounter the pose and illumination changes in uncontrolled environment. In fact, there are tight correlations for images from different views. Images of different views are the projection of the same object to different positions. Therefore local features are more useful for recognition across view than the global feature where the global structure is destructed by the projection. Thus we proposed a network which extracted the local features by $1\times1$ convolutional layers, in addition multi-hierarchical features are combined for the task of recognition. Experiments on MultiPIE dataset showed very good and stable performance of the proposed networks in a wide range of pose and illumination.

\section*{Acknowledgment}
This paper is partly supported by National Nature Science Foundation of China (No. 61502388), Ph.D. Programs Foundation of Ministry of Education of China(No. 20136102120041), the Fundamental Research Funds for the Central Universities(No. 3102015BJ(II)ZS016), the Seed Foundation of Innovation and Creation for Graduate Students in Northwestern Polytechnical University(No.Z2016021).

\bibliographystyle{IEEEbib}
\bibliography{ref_face_1.2}
\balance

\end{document}